\documentclass{article}

\usepackage[preprint]{neurips_2026}

\usepackage{comment}
\usepackage{subcaption}
\usepackage[utf8]{inputenc} 
\usepackage[T1]{fontenc}    
\usepackage{hyperref}       
\usepackage{url}            
\usepackage{booktabs}       
\usepackage{amsfonts}       
\usepackage{nicefrac}       
\usepackage{microtype}      
\usepackage{xcolor}         
\usepackage{algorithm}
\usepackage{algpseudocode}
\usepackage{microtype}
\usepackage{graphicx}
\usepackage{subcaption}
\usepackage{float}
\usepackage{booktabs} 
\usepackage{subcaption}
\usepackage{booktabs}
\usepackage{capt-of}
\usepackage{xspace}

\usepackage{hyperref}
\usepackage{caption}
\usepackage{algorithmicx}
\usepackage{multirow}
\usepackage{enumitem}
\usepackage{algorithmicx}
\usepackage{multirow}
\usepackage{enumitem}
\usepackage{amsmath}
\usepackage{amssymb}
\usepackage{mathtools}
\usepackage{wrapfig}
\usepackage{amsthm}
\usepackage{colortbl}      
\usepackage[table]{xcolor} 
\definecolor{highlightColor}{HTML}{9d2246} 
\usepackage{array}         
\usepackage{siunitx} 
\usepackage[makeroom]{cancel}
\usepackage{url}
\usepackage[utf8]{inputenc}
\usepackage[textsize=tiny]{todonotes}

\definecolor{tablecolor}{rgb}{0.8,0.8,0.8}

\newcommand{\losshat}{\hat{\ell}}

\newcommand{\vparam}{\boldsymbol{\theta}}

\newcommand{\dkl}[3]{\mathbb{D}_{\text{KL}}^{#1}(#2 \, \|\, #3)}

\newcommand\cut[1]{}

\newcommand{\squishlist}{
   \begin{list}{$\bullet$}
    { \setlength{\itemsep}{0pt}      \setlength{\parsep}{3pt}
      \setlength{\topsep}{3pt}       \setlength{\partopsep}{0pt}
      \setlength{\leftmargin}{1.5em} \setlength{\labelwidth}{1em}
      \setlength{\labelsep}{0.5em} } }

\newcommand{\squishlisttwo}{
   \begin{list}{$\bullet$}
    { \setlength{\itemsep}{0pt}    \setlength{\parsep}{0pt}
      \setlength{\topsep}{0pt}     \setlength{\partopsep}{0pt}
      \setlength{\leftmargin}{2em} \setlength{\labelwidth}{1.5em}
      \setlength{\labelsep}{0.5em} } }

\newcommand{\squishend}{
    \end{list}  }

{}
{}
{}

\newcommand{\half}{\mbox{$\frac{1}{2}$}}

\newcommand{\sqr}[1]{\left[#1\right]}

\newcommand{\myexpect}{\mathbb{E}}

\newcommand{\gauss}{\mbox{${\cal N}$}}

\newcommand{\myvec}[1]{\mbox{$\mathbf{#1}$}}
\newcommand{\myvecsym}[1]{\mbox{$\boldsymbol{#1}$}}

\newcommand{\vsigma}{\mbox{$\myvecsym{\sigma}$}}
\newcommand{\vSigma}{\mathbf{\Sigma}}

\newcommand{\vg}{\mathbf{g}}
\newcommand{\vh}{\mbox{$\myvec{h}$}}

\newcommand{\vm}{\mathbf{m}}

\newcommand{\vs}{\mbox{$\myvec{s}$}}

\newcommand{\vx}{\mbox{$\myvec{x}$}}

\newcommand{\vy}{\mbox{$\myvec{y}$}}

\newcommand{\vI}{\mbox{$\myvec{I}$}}

\newcommand{\vS}{\mbox{$\myvec{S}$}}
\newcommand{\vT}{\mathbf{T}}

\newcommand{\calD}{\mbox{${\cal D}$}}

\newcommand{\data}{\calD}

\usepackage[capitalize,noabbrev]{cleveref}
\theoremstyle{plain}

\theoremstyle{definition}

\theoremstyle{remark}

\makeatletter
\AddToHook{cmd/appendix/before}{\def\cref@section@alias{appendix}\def\cref@subsection@alias{appendix}}
\makeatother

\newcommand{\method}{CoVON\xspace}

\title{Fast and Slow Variational Continual Learning}

\definecolor{TolMutedBlue}{HTML}{332288}
\definecolor{TolMutedGreen}{HTML}{117733}
\definecolor{TolMutedPurple}{HTML}{AA4499}
\definecolor{TolMutedRed}{HTML}{BB3322}

\hypersetup{
    colorlinks=true,
    citecolor=TolMutedBlue,%
    linkcolor=TolMutedGreen,%
    urlcolor=TolMutedPurple,%
}

\author{
Subarnaduti Paul$^{1}$, Yohan Jung$^{2}$, Mohammad Emtiyaz Khan$^{3,4,5}$, 
\hspace{0.15em} \textbf{Siddharth Swaroop}$^{6}$,\\ \textbf{Thomas Möllenhoff}$^{4}$,
 \textbf{Martin Mundt}$^{1}$
 \\[0.5em]
$^1$Faculty of Mathematics and Computer Science, University of Bremen, Bremen, Germany, \\
$^2$ Department of Computer Science \& AI, Jeonbuk National University, Republic of Korea, \\
$^3$ Department of Computer Science, TU Darmstadt, Darmstadt, Germany, \\
$^4$ RIKEN Center for Advanced Intelligence
Project, Tokyo, Japan, \\
$^5$ Hessian Center for AI (hessian.AI), Darmstadt, Germany, \\
$^6$ University College London, United Kingdom. \\
{\tt\small \{spaul, mundtm\}@uni-bremen.de} 
}

\DeclareMathOperator{\tr}{tr}
\crefname{section}{Sec.}{Sec.}
\crefname{thm}{Thm.}{Theorem}
\crefname{appendix}{App.}{Appendices}
\Crefname{appendix}{App.}{Appendices}
\crefname{algorithm}{Alg.}{Algorithms}
\crefname{equation}{Eq.}{Eqs.}
\crefname{figure}{Fig.}{Figs.}
\crefname{prop}{Prop.}{Props.}
\creflabelformat{equation}{#2\textup{#1}#3}

\begin{document}

\maketitle

\begin{abstract}

Continual learning remains a major challenge for modern deep networks, partly because commonly used optimizers lack inherent mechanisms for continual adaptation. 
One such natural mechanism is ‘fast and slow adaptation’ to balance stability and plasticity. 
This mechanism has deep roots in neuroscience and biology, but there is no consensus on how to
best incorporate it in commonly used optimizers. Here, we show that this can be easily done via the VCL framework, where past posteriors are used as priors in the future. Our key idea is to incorporate slow adaptation via merging of past posteriors to slow down the drift in the knowledge as learning progresses. The merged posterior is then used as the prior in the VCL update to implement the fast-weight updates. These steps can be seamlessly implemented in the IVON
optimizer, whose form and costs are nearly identical to that of Adam. We call this new optimizer the Continual IVON (CoVON) optimizer and show that
it not only consistently improves over existing VCL optimizers, but also performs better than other weight-regularization strategies across domain-incremental learning, continual pre-training, and fine-tuning of large language models.
\end{abstract}

\section{Introduction} 

Modern deep learning systems, especially large language models (LLMs), are increasingly deployed in non-stationary environments, with data and tasks evolving over time~\citep{silver2025welcome}. 
This is the central problem in continual and lifelong learning~\citep{thrun1998lifelong,hadsell2020embracing}, where a fundamental challenge is to sequentially update these models without catastrophically interfering with the previously learned knowledge. 
Forgetting occurs partly because standard optimizers, such as SGD and Adam, are not designed to handle non-stationary settings and therefore do not have built-in mechanisms to preserve old knowledge. 
As a result, continual learning is treated as an add-on problem, typically addressed through external regularization \citep{kirkpatrick2017overcoming,zenke2017continual,Lee2017IMM}, experience or data replay \citep{rebuffi2017icarl,rolnick2019experiencereplaycontinuallearning,lopez2017gradient}, or isolation of architectural components \citep{Rusu2016,DEN,SENN}, see surveys for a full overview~\citep{Mundt2023wholistic,wang2024comprehensive}. 
This is a rather complicated way to deal with forgetting. Ideally, we want knowledge retention to be inherently rooted in the optimization process itself, such that an optimizer should naturally be able to balance effective adaptation to the current task as well as careful preservation of previous knowledge. 

One way to naturally balance such stability-plasticity is to use fast and slow adaptation. Essentially, we want to operate learning on two distinct timescales: a fast component that adapts rapidly to new data, and a slow component that consolidates and preserves previously acquired knowledge. This is rooted in Complementary Learning Systems (CLS) theory from neuroscience~\citep{CLS-Mcclelland,OReilly2003}, where the hippocampus is associated with fast adaptation and the neocortex with slow consolidation. Motivated by this view, many works for continual learning have drawn loose inspiration from CLS theory. 
However, instead of incorporating the fast and slow dynamics explicitly in the optimization of a single model, previous works use it implicitly to inform the design of structure or memory~\citep{dorovatas2026modularmemorykeycontinual}. 
For example, they focus on the interplay between generative and discriminative models (neocortex and hippocampus) with a replay mechanism~\citep{shin2017continual,vandeven2020Brainreplay,Lee2016} and the separation into disjoint sets of parameters~\citep{DualNetsfastslow,CLS-ER,RL-CLS}. Little work has been done on directly incorporating such updates into the optimizers for continual learning, and so far, there is no consensus on how to add this feature to commonly used optimizers. 

In this work, we take a different perspective and show that fast and slow adaptation can be naturally implemented within the variational continual learning (VCL) framework~\citep{nguyen2017variational}. 
This framework uses past posteriors as prior distributions to update the new posterior.
We show that slow-weight updates can be implemented via ``posterior merging'' of old posteriors, essentially slowing down the drift in the knowledge as new tasks are observed in an uncertainty-weighted manner. 
With this slow-moving prior, the fast-weight updates are implemented using the VCL objective. For this, we use a recently-proposed variational training algorithm, called {I}mproved {V}ariational {O}nline {N}ewton optimizer \citep{shen2024variational}, which uses diagonal Gaussian posteriors and whose form and costs are nearly identical to that of Adam. We call this new algorithm Continual~IVON~(\method). 
Here, we show the benefits of \method for continual learning over existing VCL and other weight-regularization strategies through various experiments across classic domain-incremental learning, continual pre-training, and fine-tuning of large language models.

\begin{figure}[t!]
    \centering    \includegraphics[width=\linewidth]{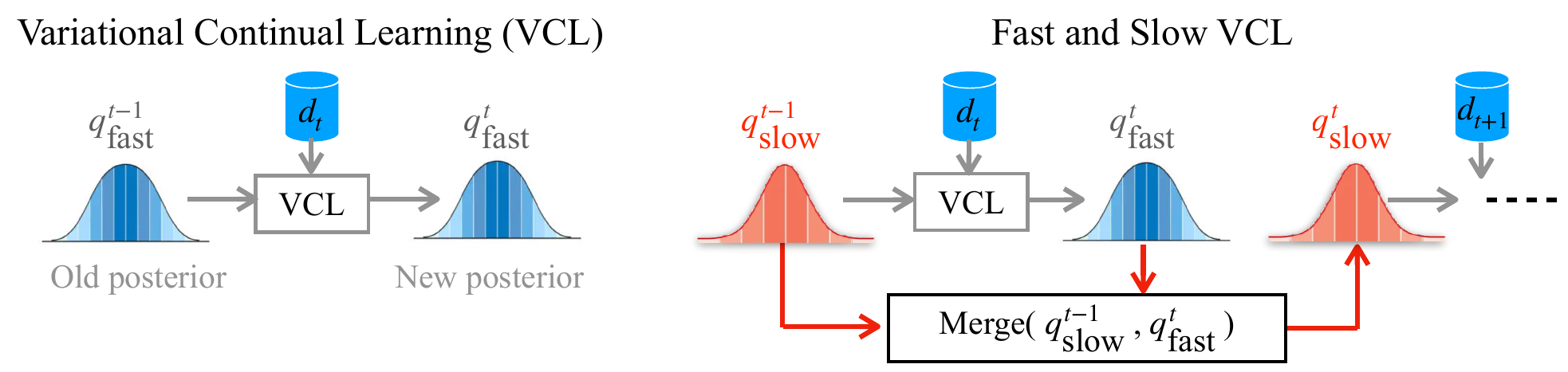} 
    \caption{CoVON balances stability and plasticity in Variational Continual Learning through fast and slow adaptation steps in a single model, where the VCL block denotes the variational update in \Cref{eq:vcl}. The slow-moving posterior (red) slowly consolidates information from the previous posterior and the new fast posterior (blue), merging the models together. The fast posterior is trained on the current data using the previous slow posterior as a new prior in a Bayesian continual update. 
    }
    \label{fig:fig1}
\end{figure}

\section{Background and Related Work}
Continual learning algorithms for neural networks must balance complementary properties:  \emph{plasticity}, the ability to adapt rapidly to new tasks without 
losing the capacity to learn; and \emph{stability}, the ability to preserve 
knowledge acquired in previous tasks without catastrophic forgetting.
A natural mechanism to balance this is `fast and slow adaptation', where plasticity is maintained through fast-moving elements, and stability is maintained in the slow-moving parts. 

However, most popular continual learning methods attempt to balance these properties using external mechanisms added on top of the standard training pipeline. 
These include regularization-based, replay/memory-based, and architecture-based approaches, see surveys for a full overview~\citep{Mundt2023wholistic,wang2024comprehensive}. 
These approaches attempt to correct for a broader limitation of current optimization algorithms, such as SGD and AdamW \citep{LoHu17}: while effective to learn the task at hand, they do not explicitly preserve knowledge over time, offering limited balance to the stability-plasticity tradeoff, and eventually succumbing to catastrophic forgetting \citep{mirzadeh2020understanding}. 
Other recent works introduce simple methods that do not have external mechanisms and are similar to existing deep-learning optimizers: \citet{li2025fishers} proposes reusing AdamW's scale vector for continual learning, but this can be inaccurate for larger batch sizes. Utility-Perturbed Gradient Descent (UPGD)~\citep{elsayed2024addressing} introduces a continual optimizer, but it relies on second-derivative computations, which are difficult to scale to modern, larger-scale neural networks. 
In this work, we introduce a simple standard training pipeline that computes second-order estimates in an online fashion within the optimization steps without significant overhead, enabling continuous adaptation at scale. 

Instead of finding external mechanisms that may or may not balance fast adaptation and slow consolidation well, we can instead directly use learning dynamics that operate at the two different paces. 
This has been studied in neuroscience and is used to motivate machine learning algorithms. 
The neuroscience theory of complementary learning systems (CLS) by \citet{CLS-Mcclelland} argues that the human brain relies on fast and slow learning through the hippocampus and neocortex. 
Continual learning works inspired by this CLS view, however, use heuristic methods. 
They either explicitly use two sets of weights or dual-system mechanisms; these include CLS-ER \citep{CLS-ER}, Dual-Nets \citep{DualNetsfastslow}, RL-based CLS \citep{RL-CLS}, and MERLIN \citep{MERLIN}. 
In contrast, we naturally incorporate fast and slow updates within the optimization process of a single model. 

Another line of work is inspired by biological views on fast and slow updates of synapses \citep{hinton1987}, using fast weights to deblur compressed memories stored in slow-moving weights. This perspective was later revisited to speed up optimization in deep learning~\citep{ZhLu19}, for sequence modeling~\citep{hinton-DL}, for meta-learning~\citep{finn2017model,nichol2018first}, and recently, for continual learning~\citep{abbes2025revisiting}. 
Our work presents a variational Bayesian formulation of such views, along with a scalable implementation for Gaussian posteriors, which gives new curvature-weighted variants of these methods.

Here, our goal is to show how we can naturally implement fast and slow adaptation in the VCL framework \citep{nguyen2017variational}. 
Unlike standard training methods that aim to find a parameter $\vparam$ by minimizing a loss function {$\ell_{t}(\vparam) = \sum_{i \in \data_{t}} \ell_i(\vparam)/|\data_t|$}, VCL aims to find a distribution $q(\vparam)$ by optimizing the Kullback-Leibler divergence objective given a past posterior $q_{t-1}$ as prior at task $t$.
The optimization proceeds as follows, 
\begin{equation}
   q_{t}(\vparam) \gets \arg \min_{q \in \mathcal{Q}} ~ \lambda_{t} \mathbb{E}_{q(\text{$\vparam$})}[\ell_{t}(\vparam)] + \dkl{}{q(\vparam)}{q_{t-1}(\vparam)}. 
   \label{eq:vcl}
\end{equation}
It can be shown that by making certain approximations, solutions of standard 
optimizers can be recovered as special cases of this objective. 
Contrary to the original VCL work, we introduce a scaling factor $\lambda_{t} > 0$, which recovers VCL when set to the number of data examples $|\data_t|$. 
Although VCL can perform decently in continual learning due to the Bayesian update structure, it does not have a sense of fast and slow updating: only one set of weights are updated using the same optimization procedure. 
This means that there is no explicit tradeoff between fast adaptation and slow consolidation. 
\citet{nguyen2017variational} argued that VCL satisfies this automatically due to the Bayesian updates; however, in practice, VCL's performance can deteriorate rapidly due to approximations required for neural networks (such as mean-field Gaussian approximations). 
In our method, we will introduce a separate slow update to naturally include fast and slow dynamics. We next describe how we build on VCL. 
\algnewcommand{\Init}[1]{\Statex \hspace*{-\algorithmicindent}\textbf{Init:} #1}
\algnewcommand{\Optional}[1]{\Statex \hspace*{-\algorithmicindent}\textbf{Optional:} #1}
\algnewcommand{\Set}[1]{\State \textbf{Set:} #1}
\algrenewcommand\algorithmiccomment[1]{\hfill\textcolor{blue}{// #1}}

\section{Fast and Slow VCL via Continual Variational Online Newton (\method)} 

We now show how we naturally implement fast and slow mechanisms within the VCL framework, where the fast adaptation controls the plasticity on the new task and the slow adaptation consolidates the prior knowledge acquired from the past tasks.
We first describe the general framework 
in \Cref{sec:assumptions} and then present two scalable implementations using diagonal Gaussian posteriors: a naive approximation that uses AdamW, and then our full CoVON algorithm (that uses IVON). 

\subsection{Fast and Slow Variational Continual Learning}\label{sec:assumptions}

We propose to generalize VCL into a two-stage fast and slow update (see also \Cref{fig:fig1}), where (1) a fast adapted per-task posterior is learned using \Cref{eq:vcl}, and then (2) it is consolidated into a slow-moving outer posterior obtained through merging of past information with new information. 
For fast adaptation, we use \Cref{eq:vcl}, optimizing for \smash{$q_{t}^{\textrm{fast}}$} while setting the prior to be \smash{$q_{t-1}^{\textrm{slow}}$}.  

\textbf{Slow Adaptation: }
After learning the fast posterior $q_t^{\textrm{fast}}$ 
on the current task, we consolidate it into a slow-moving outer posterior by merging it with the previous slow-moving posterior \smash{$q_{t-1}^{\textrm{slow}}$} with the newly learned posterior \smash{$q_{t}^{\textrm{fast}}$}:

\begin{equation}
   q_{t}^{\textrm{slow}}(\vparam) \propto q_{t-1}^{\textrm{slow}}(\vparam) \left(\frac{q_{t}^{\textrm{fast}}(\vparam)}{q_{t-1}^{\textrm{slow}}(\vparam)} \right)^\gamma .
   \label{eq:post-merge}
\end{equation}
This choice of update enables us to strike a better balance between the past and the future, in a similar fashion as other online learning methods. This can be shown by noting a property of the solutions obtained by solving \cref{eq:vcl}. Essentially, as shown by \citet[Eq. 4]{khan2025knowledge} (and many other works before), a solution of \cref{eq:vcl} has the following form in terms of the \emph{site} function $\losshat_{t}(\vparam)$, 
\begin{equation}
    q_t^{\textrm{fast}}(\vparam) \propto q^{\textrm{slow}}_{t-1}(\vparam) \exp \sqr{- \lambda_t \losshat_t(\vparam)}, 
\end{equation}
where the {site} function $\losshat_{t}(\vparam)$ summarizes the effect of the task loss $\ell_t$ on our learned distribution $q_t^{\textrm{fast}}$ (see \Cref{app:derivation-vclgeneral} for derivation).
Substituting this in \cref{eq:post-merge}, we can write the slow update posterior as
\begin{equation}
    q_t^{\textrm{slow}}(\vparam) \propto q_{t-1}^{\textrm{slow}}(\vparam)  \exp \sqr{- {\color{red} \gamma} \lambda_t \losshat_t(\vparam)} 
    \,\,\, \propto q_0^{\textrm{slow}}(\vparam) \prod_{i=1}^t \exp \sqr{- {\color{red} \gamma} \lambda_i \losshat_i(\vparam)}. 
\end{equation}
The second expression is obtained by recursively applying the form. The expression shows that the slow update is down-weighting all the past site functions (by $\gamma$ highlighted in red). This has a net effect to slow down the learning, essentially reversing the effect of a larger $\lambda_t$ used in the fast update. Such exponentially weighted aggregation \citep{vovk1990aggregating} is a common strategy in online learning and can recover optimal guarantees by changing the posterior form~\citep{hoeven2018many}.
 
Our updates also have a similar property, where changing the form of the posterior can recover previous proposals for fast and slow adaptation. 
For instance, for simple isotropic Gaussian distributions with fixed-variance, \Cref{eq:post-merge} recovers shortest descent~\citep{hinton1987}, the lookahead optimizer~\citep{ZhLu19} and first-order meta-learning methods such as Reptile~\citep{nichol2018first}. We would expect that using more expressive posteriors improves the quality of this knowledge consolidation step. In the next subsection, we will see that using Gaussians with diagonal variance, one obtains curvature or uncertainty-weighted variants of these methods.

\paragraph{Fast adaptation.}Given the slow posterior $q_{t-1}^{\textrm{slow}}$ 
as a stabilizing prior, we optimize for the fast posterior $q_t^{\textrm{fast}}$ 
by solving \Cref{eq:vcl} with $q_{t-1}^{\textrm{slow}}$ in place of $q_{t-1}$:
\begin{equation}
   q_{t}^{\textrm{fast}}(\vparam) \gets \arg \min_{q \in \mathcal{Q}} ~ 
   \lambda_{t} \mathbb{E}_{q(\vparam)}[\ell_{t}(\vparam)] + 
   \dkl{}{q(\vparam)}{q_{t-1}^{\textrm{slow}}(\vparam)}.
\end{equation}
Here $\lambda_t > 0$ acts as a tunable learning-rate-like parameter controlling 
the balance between fitting the new task and retaining the slow posterior. While 
a natural default is $\lambda_t = |\mathcal{D}_t|$, this is not a natural choice in online settings~\citep{hoeven2018many,cherief2019generalization}, and we 
therefore treat it as a hyperparameter. 

\subsection{Diagonal-Gaussian Implementation and the Proposed \method Method}
\label{sec:derivation}

We now derive two practical variants of the VCL algorithm. We start with a simple variant of VCL that reuses curvature estimates obtained from training with AdamW, and then present our more faithful CoVON algorithm, which also uses the fast and slow model merging updates.
  
To derive the Adam variant, we first note that any Gaussian solution of form $q_t(\vparam)=\gauss(\vparam\,|\,\vm_t, \vSigma_t)$ to \Cref{eq:vcl} satisfies the following optimality conditions, 
\begin{align}
   &\vS_{t-1} (\vm_{t} - \vm_{t-1}) = -\lambda_t \mathbb{E}_{q_{t}(\text{$\vparam$})}[\nabla \ell_{t}(\vparam)],
   & \vS_{t} = \vS_{t-1} + \lambda_t \mathbb{E}_{q_{t}(\text{$\vparam$})}[\nabla^2 \ell_{t}(\vparam)], \label{eq:prec_update}
\end{align} 
where $\vS_t = \vSigma_t^{-1}$ is the precision matrix (inverse covariance). This is obtained by setting the gradient with respect to mean and precision in \Cref{eq:vcl} to zero (we have dropped superscripts `fast' and `slow'), see \citep[Sec.~3.1]{KhRu23}. We present a self-contained derivation of this in \Cref{app:derivation-optcond}. 

Performing an approximation to the expectation using a single sample at the distribution's mean in \Cref{eq:prec_update} leads to the following approximate VCL update as shown below:

\begin{minipage}{0.52\linewidth}
\vspace{-0.3cm}
\begin{equation}
  \vparam_{t} \gets \arg \min_{\text{$\vparam$}} \, \ell_{t}(\vparam) + \frac{1}{2\lambda_{t}} \| \vparam - \vparam_{t-1} \|_{\text{$\vS_{t-1}$}}^2
  \label{eq:ewc1}
\end{equation}
\end{minipage}\hfill
\begin{minipage}{0.47\linewidth}
\vspace{-0.3cm}
\vspace{0.23\baselineskip}
\begin{equation}
  \vS_{t} \gets \vS_{t-1} + \lambda_{t} \nabla^2 \ell_{t}(\vparam_{t}).
  \label{eq:ewc2}
\end{equation}
\end{minipage}

To derive this, we have rewritten the first optimality condition as an optimization problem and switched the notation from $\vm$ to $\vparam$ to highlight the approximation. This yields the precision for the next task \smash{$\vS_{t} = \vS_0 + \sum_{k=1}^{t} \lambda_k \nabla^2 \ell_{k}(\vparam_{k})$} as an accumulation of past Hessians, where $\vS_0$ is a prior chosen for the first task, typically $\vS_0 = \lambda_1 \delta \vI$ corresponds to weight-decay with strength $\delta > 0$.  

In practice, one further approximates the Hessian using the squared gradients:
\begin{align}
&\nabla^2 \ell_t(\vparam) 
\approx \text{diag}(\vh), 
&\vh = 
\mathbb{E}_{(\text{\vx}, \text{\vy}) \sim \mathcal{D}_t} 
\big[(\nabla_{\vparam} \mathcal{L}(f(\vparam, \vx), \vy))^2\big].
\label{eq:sqgrad}
\end{align}
The resulting update affects \Cref{eq:ewc1} by changing the regularizer term to become $\smash{\half \sum_{t=1}^T (\vparam - \vparam_t)^\top \mathrm{diag}(\vh_t) (\vparam - \vparam_t)}$ for each task $t$. This is the regularizer used in classical works such as  Elastic Weight Consolidation (EWC) \citep{kirkpatrick2017overcoming} and EWC* \citep{huszar2018note}. Both methods approximate posterior precision through Hessians at a single point, but do so only after training each task, requiring an extremely expensive additional full pass over the dataset to compute \Cref{eq:ewc2} together with \Cref{eq:sqgrad} in a post-hoc fashion. As a cheap approximation, we propose to recycle the squared-gradient accumulator in AdamW, and we refer to this method as Ada-Reg. The pseudo-code of our EWC-like and Ada-Reg methods are provided in \Cref{app:adareg}. 

\definecolor{lightgreen}{rgb}{0.56,0.93,0.56}
\begin{algorithm*}[t]
\caption{\method: Fast and Slow Variational Continual Learning}\label{alg:ivon_vcl}
\begin{algorithmic}[1]
\Init{
$\vm \gets (\text{NN-weight init}), \;
\vm_0 \gets 0, \;
\vs_0 \gets \text{prior-precision}$
}
\Init $\beta_1, \beta_2 \in (0,1)$,  $\lambda_t$ for each task $t$, Hessian init $\vh_0 = h_0 \cdot \mathbf{1}$, \text{learning rates} $\alpha$, $\gamma$ 
\For{$t=1\hdots T$} 
\State $\vg \gets 0$, $\vh \gets h_0$,  \quad 

\For{$i=1,2,\hdots$}   
\State Sample $\vparam \sim \gauss(\vparam \, | \, \vm, \vsigma^2)$ with $\vsigma^2= 1 / (\lambda_{t} \vh + \vs_{t-1})$
\State $\widehat \vg \gets \widehat \nabla \ell_{t}(\vparam)  $
\State $\widehat \vh \gets \widehat \vg (\vparam - \vm) / \vsigma^2$ \hspace{2.5in}\rlap{\smash{$\left.\begin{array}{@{}c@{}}\\{}\\{}\\{}\\{}\\{}\\{}\\{}\end{array}\color{highlightColor}\right\}%
          \color{highlightColor}\begin{tabular}{l} Fast VCL Update\\ using IVON \& \\ Slow-Updated Prior \end{tabular}$}}
\State $\vg \gets \beta_1 \vg + (1-\beta_1) \widehat \vg$, \quad $\bar \vg \gets \vg / (1 - \beta_1^i)$
\State $\vh \gets \beta_2 \vh + (1-\beta_2) \widehat \vh + \half (1 - \beta_2)^2 (\vh - \widehat \vh)^2 / (\vh + \lambda_{t}^{-1} \vs_{t-1})$ 
\State $
\vm \gets \vm - \alpha (\bar \vg + \textcolor{highlightColor}{\displaystyle \lambda_{t}^{-1} \vs_{t-1} (\vm - \vm_{t-1})})/( \vh + \lambda_{t}^{-1} \vs_{t-1}) 
$
\EndFor 
\State \textcolor{highlightColor}{$\vs_{t} \gets (1 - \gamma) \vs_{t-1} + \gamma (\vs_{t-1} + \lambda_{t} \vh)$} 
\State \textcolor{highlightColor}{$\vm_{t} \gets \left[(1 - \gamma) \vs_{t-1} \vm_{t-1} + \gamma (\vs_{t-1} + \lambda_{t} \vh) \vm \right] / \vs_{t}$} \hspace{0.975in} \rlap{\smash{$\left.\begin{array}{@{}c@{}}\\{}\\{}\\{}\end{array}\color{highlightColor}\right\}%
          \color{highlightColor}\begin{tabular}{l} Slow Update\end{tabular}$}} 
\State \textcolor{highlightColor}{$\vm \gets \vm_{t}$}
\EndFor
\end{algorithmic}  
\label{alg:vcl-ivon}
\end{algorithm*} 
Instead of approximating the expectation with a single sample at the mean, our method \method uses the Improved Variational Online Newton method (IVON), see \citet[Alg.~1]{shen2024variational} to directly solve \Cref{eq:prec_update}. \method also uses the additional proposed slow update. Algorithm~\ref{alg:ivon_vcl} summarizes the  final method, and in the following, we will describe the individual components.

\paragraph{\method's Inner Loop (Fast Adaptation).} CoVON closely follows the Ada-Reg pseudo-code (see \Cref{app:adareg}), but solves the variational problem \Cref{eq:vcl}. This is achieved by sampling parameters from the current posterior $q_t$ in \text{line 4} of Algorithm~\ref{alg:ivon_vcl},
and evaluating gradients at the sampled parameters in \text{line 5}. This stochastic parameter sampling allows \method to propagate posterior uncertainty through the optimization trajectory and estimate curvature information rather than collapsing prematurely to a single solution.  This noise is also crucial for controlling the model’s plasticity, with noisy weights corresponding to plastic ones and static weights to stable ones.  
The noise is also used to estimate the Hessian $\vh$ in \cref{eq:sqgrad} via the reparametrization trick in line~6, giving curvature estimates for free. 
To retain the posterior estimate from the previous tasks, which is essential for continual learning, we augment each parameter update step with a quadratic regularizer centered at the previous posterior mean $\vm_{t-1}$ in line 9. 
The regularizer is also scaled by task-specific precision $\vs_{t-1}$, which captures curvature information updated online across tasks. Together, this implements the KL-term in \Cref{eq:vcl}.

\paragraph{\method's Outer Loop (Slow Consolidation).} Instead of simply accumulating the Hessians in the outer loop as suggested by \cref{eq:prec_update} and done in Ada-Reg (\Cref{app:adareg}), CoVON now uses the merging from \Cref{eq:post-merge}, which in our Gaussian case takes the following form: 
\begin{align}
   \vs_{t} \gets (1 - \gamma) \vs_{t-1} + \gamma (\vs_{t-1} + \lambda_{t} \vh), \quad 
     \vm_{t} \gets \left[(1 - \gamma) \vs_{t-1} \vm_{t-1} + \gamma (\vs_{t-1} + \lambda_{t} \vh) \vm \right] / \vs_{t},
     \label{eq:mm2}
\end{align}

where $\vh$ is the online Hessian estimated during the inner loop. This avoids any expensive post-hoc Fisher or Hessian evaluations as in EWC. This can also be well achieved without any task boundaries, where one may just need to set a threshold in the number of optimization steps or track the loss curve to determine the point of convergence as an accumulation point. 

For $\gamma=1$, the above simplifies to \Cref{eq:prec_update} which we use in Ada-Reg and EWC.
A choice of $\gamma < 1$ favors retaining old information, and consolidates the knowledge from the inner fast adaptation into the slow-moving outer posterior. Such ideas have existed for a long time, see \citet{hinton1987}. The idea is that the fast moving set of weights in the inner loop attends to the new tasks and \emph{deblurs} old memories stored in the slow moving weights (outer loop). Knowledge in the fast moving weights then gets merged or consolidated into a set of slow moving weights in the outer loop that stores the long term memories. The updates  \Cref{eq:mm2} yield a curvature-weighted uncertainty-aware generalization of such fast-and-slow learning algorithms~\citep{hinton1987,nichol2018first,ZhLu19}. Finally, we note that a choice of $\gamma > 1$ is also possible and corresponds to an overrelaxation which is frequently used in optimization~\citep{EcBe92}, but appears less useful for continual learning. 
The updates in \Cref{eq:mm2} are also an instance of uncertainty-guided model merging~\citep{daheim2023model}, which augments regular model-merging with a weighting. The intuition is that weights with high uncertainty (noisy, plastic weights which can take on any value) should move more in the model merging than weights which have little uncertainty and are static. 

\section{Experiments: Fast \& Slow Variational Continual Learning with \method}
We now empirically corroborate our fast and slow VCL framework and its instantiation through \method. We begin by revisiting VCL through the lens of fast and slow adaptation in Section \ref{sec:exp_revVCL}, demonstrating that \method improves over existing VCL optimizers and ablating key design choices in the process. 
We then run prevalent domain incremental benchmarks in Section \ref{sec:exp_context}, demonstrating that \method performs favorably against other (non-VCL-based) SOTA CL algorithms. Finally, we validate \method's scalability and robustness in more challenging and realistic scenarios. Here, we highlight \method's aptitude for continual pre-training across multilingual data (\Cref{sec:exp_pretraining}) in LLMs and knowledge preservation in fine-tuning on common reasoning benchmark (\Cref{sec:exp_reasoning}). 

\begin{figure}[t]
\begin{subfigure}[t]{0.5\linewidth}
    \vspace{0pt}
    \centering
    \includegraphics[width=0.95\linewidth]{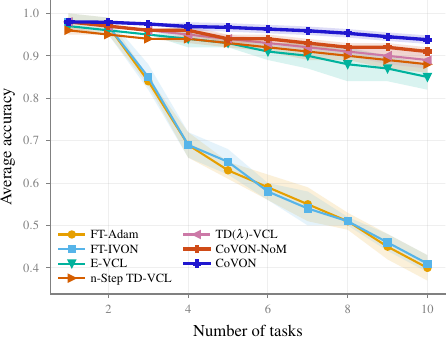}
    \phantomcaption
    \label{fig:TD-VCl}
\end{subfigure}%
\hspace{0.01\linewidth}%
\begin{subfigure}[t]{0.48\linewidth}
    \vspace{15pt}
    \centering
    \scriptsize
    \setlength{\tabcolsep}{5pt}
    \resizebox{\linewidth}{!}{%
\begin{tabular}{l l l c c}
\toprule
\textbf{Update} & \textbf{Prior} & \textbf{Outer loop} & $\boldsymbol{\gamma}$ & \textbf{Avg Acc} \\
\midrule
\multicolumn{5}{l}{\textcolor{gray}{\textit{Fast only}}} \\
Fast        & w/o & $-$      & $=1$             & $40.23\pm 2.98$ \\
Fast        & w/  & $-$      & $=1$           & $77.03\pm1.03$ \\
\midrule
\multicolumn{5}{l}{\textcolor{gray}{\textit{Fast \& slow}}} \\
Fast \& slow & w/ & EMA      & $=1$            & $89.26\pm0.24$ \\
Fast \& slow & w/ & EMA      & $<1$ & $89.56\pm0.25$ \\
Fast \& slow & w/ & Full VCL & $=1$  & $90.32\pm0.27$ \\
\rowcolor[gray]{0.92}
Fast \& slow & w/ & Full VCL & $<1$   & \textbf{$92.12\pm0.35$} \\
\bottomrule
\end{tabular}}
    \phantomcaption
    \label{tab:ablation_mean}
\end{subfigure}
\caption{(a) Comparison of \method against recent VCL approaches on a Sequential Permuted MNIST-Hard benchmark. Here, \method-NoM denotes the variant without posterior merging (this is then the same as VCL using natural-gradient updates), whereas \method includes this slow update, as shown in \Cref{eq:mm2}. CoVON-NoM already outperforms the strongest baseline TD-VCL, and the merging-based update further improves the model performance, demonstrating the importance of the interplay between fast and slow components. (b) Ablation study of the fast and slow components of \method. Considering only fast IVON-like updates unsurprisingly leads to catastrophic forgetting, which is partially alleviated by adding a prior. On the contrary, using both fast and slow mechanisms offers a significant improvement, where the full fast and slow VCL mechanism of \method outperforms a curvature-unaware EMA-based merging of past posteriors.}
\label{fig:VCL-Base}
\end{figure}
\begin{table*}[t]
\centering
\scriptsize
\caption{Comparison of \method against non-VCL SOTA baselines on domain-incremental benchmarks. We report the average accuracy $A_T$ and the average forgetting $F_T$ on DomainNet, CDDB-Hard, and CORe50 under two backbone settings: a standard ViT-B/16 and CLIP with a pretrained ViT-B/16 vision encoder. Best results per section are highlighted in \textbf{bold}. \method consistently improves both retention and final accuracy over prior methods and further outperforms the no-merging variant (\method-NoM), demonstrating the benefit of slow-merged posterior consolidation. } 
\setlength{\tabcolsep}{5pt}
\renewcommand{\arraystretch}{1.2}
\resizebox{.95\textwidth}{!}{%
\begin{tabular}{l|cc|cc|c}
\toprule
\multirow{2}{*}{\textbf{Continual learning Methods}}
  & \multicolumn{2}{c|}{\textbf{DomainNet}}
  & \multicolumn{2}{c|}{\textbf{CDDB-Hard}}
  & \textbf{CORe50} \\
\cmidrule(lr){2-3}\cmidrule(lr){4-5}\cmidrule(lr){6-6}
  & $A_{T}$\,($\uparrow$) & $F_{T}$\,($\downarrow$)
  & $A_{T}$\,($\uparrow$) & $F_{T}$\,($\downarrow$)
  & $A_{T}$\,($\uparrow$) \\
\midrule
\multicolumn{6}{c}{\textbf{Model: Vision Transformer (ViT-B/16)}} \\
\midrule
DyTox~\cite{dytoxarthur}
  & 62.94          & --
  & 86.21          & $-1.55$
  & 79.21 \\
\midrule

EWC~\citep{kirkpatrick2017overcoming}
  & $47.62{\pm0.9}$ & $-1.93$
  & $50.59{\pm1.1}$ & $-27.62$
  & $74.82{\pm1.2}$ \\

EWC*~\citep{huszar2018note}
  & $51.01{\pm0.8}$ & $-1.79$
  & $70.94{\pm0.9}$ & $-1.54$
  & $85.35{\pm1.2}$ \\

LwF~\citep{lwf}
  & $45.01{\pm0.4}$ & $-1.99$
  & $60.94{\pm0.3}$ & $-16.24$
  & $75.35{\pm0.3}$ \\

L2P~\citep{L2P}
  & $40.15{\pm0.0}$ & $-2.25$
  & $61.08{\pm0.0}$ & $-9.23$
  & $78.45{\pm0.0}$ \\

DualPrompt~\citep{dualprompt}
  & $43.79{\pm0.03}$ & $-2.03$
  & $64.80{\pm0.01}$ & $-8.74$
  & $80.68{\pm0.02}$ \\

CODA-P~\citep{coda-P}
  & $54.29{\pm0.20}$ & $-1.34$
  & $73.15{\pm0.10}$ & $-0.99$
  & $85.31{\pm0.10}$ \\

S-iPrompts~\citep{S-Prompt}
  & $50.62{\pm0.05}$ & $-2.85$
  & $74.51{\pm0.02}$ & $-1.30$
  & $83.50{\pm0.01}$ \\

PINA~\citep{PINA}
  & $54.86{\pm0.01}$ & $-2.24$
  & $77.35{\pm0.01}$ & $-0.98$
  & $87.26{\pm0.02}$ \\

\rowcolor{gray!15}
\textbf{\method-NoM}
  & $\mathbf{58.96}{\pm0.32}$ & $\mathbf{-1.86}$
  & $\mathbf{80.11}{\pm0.45}$ & $\mathbf{-0.67}$
  & $\mathbf{93.84}{\pm0.34}$ \\

\rowcolor{gray!15}
\textbf{\method}
  & $\mathbf{59.89}{\pm0.08}$             & $\mathbf{-1.61}$
  & $\mathbf{82.24}{\pm0.12}$ & $\mathbf{-0.47}$
  & $\mathbf{95.14}{\pm0.09}$ \\

\midrule
\multicolumn{6}{c}{\textbf{Model: CLIP with pretrained ViT-B/16 as vision encoder}} \\
\midrule

PINA~\citep{PINA}
  & $66.00{\pm0.2}$ & $-1.59$
  & $85.71{\pm0.0}$ & $-1.21$
  & $87.38{\pm0.1}$ \\

MoP-CLIP~\citep{Mono-CLIP}
  & $69.70{\pm0.2}$ & $-0.76$
  & $88.65{\pm0.2}$ & $-0.69$
  & $92.29{\pm0.2}$ \\

S-liPrompts~\citep{S-Prompt}
  & $68.55{\pm0.1}$ & $-1.64$
  & $86.08{\pm0.1}$ & $-1.12$
  & $90.10{\pm0.1}$ \\

\rowcolor{gray!15}
\textbf{\method}
  & $\mathbf{70.59}{\pm0.2}$ & $\mathbf{-0.53}$
  & $\mathbf{89.10}{\pm0.1}$ & $\mathbf{-0.56}$
  & $\mathbf{92.64}{\pm0.3}$ \\

\bottomrule
\end{tabular}%
}
\label{tab:main_results}
\end{table*}

\subsection{Revisiting Variational Continual Learning With Fast \& Slow Updates}\label{sec:exp_revVCL}

We first compare \method with recent VCL approaches \citep{TDVCL,EVCL} on the Permuted MNIST benchmark across 10 tasks in \Cref{fig:VCL-Base}. Although initially simplistic, we consider this similar experimental setup to the baselines \citep{TDVCL} in order to ensure comparability and avoid directly running into known scalability issues for select methods~\citep{nguyen2017variational}. 

From the figure, we can observe that \method already outperforms the strongest VCL variant (TD-VCL) \citep{TDVCL} without a slow-update merging mechanism ($\gamma < 1$, hence referred to as CoVON-NoM (CoVON No Merge)). Although \method is easy to implement in an Adam-style optimizer, it is further evident that both Adam and IVON have not been designed for such an adaptive scenario and both forget catastrophically. \method's approximately 2\% observed gain in final accuracy over the best VCL methods is further substantiated by the inclusion of the slow consolidation mechanism (\method), which merges past posteriors, boosting the final accuracy to 92.12\%. 

To shed more light on these observed improvements, we ablate the role of the interplay between the fast-adapted new posterior and the slow-consolidated old posterior in \Cref{tab:ablation_mean}. Considering only fast updates, i.e., a standard IVON optimizer, unsurprisingly yields catastrophic forgetting (38\%). On the contrary, including the posterior as the prior in standard VCL alleviates some forgetting (77\%). The clearest improvement appears when fast and slow updates are coupled through posterior merging. We ablate two variants of this slow consolidation. The first, which is the \method update in \Cref{eq:mm2}, follows the full VCL formulation and incorporates precision information through $S_t$. The second is a standard EMA-style merge, which performs a simpler weighted averaging update without the precision information ($\vm_t \gets (1 - \gamma)\vm_{t-1} + \gamma \vm$). 
The ablation results show that the slow merging step benefits from accumulated curvature information, with precision-aware consolidation consistently outperforming the simple EMA-style merge.

\subsection{Contextualizing \method In Domain Incremental Settings}\label{sec:exp_context}
We now validate \method on a set of more challenging domain-incremental learning (DIL) benchmarks without access to the task identifiers or replay buffers. Here, we further compare \method to non-VCL SOTA CL methods, in order to further contextualize \method in the broader literature landscape. To this end, we consider three standard benchmarks: CDDB~\citep{CDDB}, CORe50~\citep{core50}, and DomainNet~\citep{DomainNet}. CDDB is a continual deepfake detection benchmark, where we adopt the most demanding ``hard'' track, consisting of {$\sim$}27k images across five generative models. CORe50~\citep{core50} evaluates continual object recognition over 50 categories and 11 domains, where train sequentially on 8 domains (120k images) and test on 3 unseen domains. DomainNet~\citep{DomainNet} is a large-scale benchmark with 345 classes and $\sim$600k images spanning six diverse domains. Regarding the methods, we compare our approach against EWC \citep{kirkpatrick2017overcoming}, EWC-Corrected \citep{huszar2018note}, L2P \citep{L2P}, and LwF \citep{lwf}, as well as several exemplar-free SOTA DIL approaches, including S-Prompt \citep{S-Prompt}, PINA \citep{PINA}, CODA-P \citep{coda-P}, and our own degenerate case of Ada-Reg. We follow the same experimental protocol as our baselines, adopting a ViT-B/16 backbone pretrained on ImageNet-1k. Further experimental details are in \cref{app:exp_details}.

Table~\ref{tab:main_results} reports the average accuracy ($A_T$) and forgetting ($F_T$) across tasks at the end of training. We can observe that \method consistently strikes a strong balance between stability and plasticity across all exemplar-free domain-incremental settings. On DomainNet, it achieves an average accuracy of 59.89\%, outperforming the strongest baseline (PINA, 54.86\%) by 4.1\%, while maintaining a low forgetting value ($-1.86$). Similarly strong performance is visible for CDDB-Hard and Core50, where the latter features the largest observed gains. Changing the backbone to CLIP improves overall performance, but does not change these observed trends for \method. We attribute this to \method's fast and slow adaptation with online curvature estimates, which regularize the fast-adapted posterior mean against excessive drift under changing task distributions. As in the VCL comparison, the slow posterior-merging update can again be empirically observed to provide additional benefit.

Finally, we note that \method further reduces the cost of the computation over traditional regularization methods such as EWC and EWC$^*$. On Core50, EWC's post-hoc estimation increases computational cost by more than 200\% at task transitions, whereas \method incurs no substantial overhead beyond standard AdamW training. We provide a detailed cost comparison in \Cref{fig:cost_comp}. Overall, both \method's efficacy in balancing stability-plasticity and their computational efficiency make it an ideal contender not only for domain-incremental learning, but also for modern fine-tuning of larger models. 

\begin{figure*}[t]
    \centering
   \includegraphics[width=0.49\textwidth]{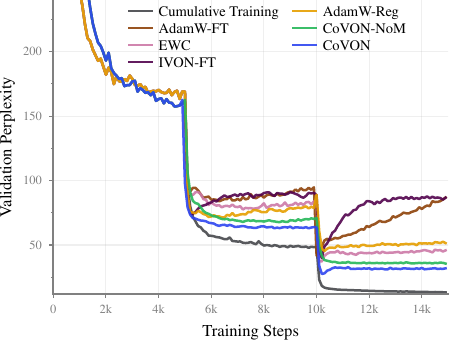}
   \includegraphics[width=0.49\textwidth ]{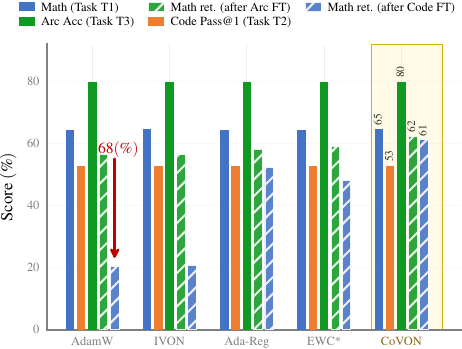}
   
    \caption{We show that a fast and slow VCL realization through \method can effectively scale to LLMs, demonstrating a strong stability-plasticity trade-off across diverse continual language modeling settings. \textbf{(a)} On continual multilingual pretraining from English to German to French (transitions after every 5k iterations), \method maintains low perplexity on previously seen languages while adapting effectively to each new language, exhibiting less forgetting than AdamW and IVON. \textbf{(b)} On continual finetuning across sequential reasoning tasks for GSM8K mathematics, Alpaca-Code, and the ARC-Challenge, \method retains approximately 95\% of mathematical reasoning accuracy after finetuning on coding and commonsense distributions (visualized as shaded bars with the color of the current fine-tuning task), while achieving plasticity on each new task comparable to AdamW.}
    \label{fig:LLM_task}
\end{figure*}

\subsection{Precision-Weighted Continual Pretraining on Multilingual Data}\label{sec:exp_pretraining}
We continually pretrain a GPT-based language model (125M parameters) across three languages in sequence: English $\rightarrow$ German $\rightarrow$ French, transitioning to a new language every 5k iterations after observing convergence. 
To assess model stability under shifting language distributions, we report the average validation perplexity across all languages seen throughout the training (\Cref{fig:LLM_task}), measuring the model's ability to acquire new linguistic knowledge, while retaining capabilities in learned languages. 

\method maintains a consistently stable trajectory throughout, achieving the lowest averaged perplexity of 31.89 by the end of training. A similar trend is observed for the variant without the slow-update posterior merging, which prevents the fast-adapted weights from overwriting old linguistic ability as each new language is introduced. In contrast, AdamW-FT and IVON-FT exhibit very high forgetting (perplexity of approx. 85), especially when French is introduced, reflecting the fact that these optimizers were designed for static learning purposes. EWC and the degenerate Ada-Reg implementation fall in between these results, alleviating some forgetting, but falling behind either \method variant. This highlights that both the variational Bayesian perspective and the addition of the curvature-weighted posterior merging in fast and slow VCL are relevant for continual adaptation at a larger scale. We hypothesize that the remaining gap to the cumulative training baseline, i.e., the baseline where all data is stored over time, is a reflection of the lack of an explicit buffer for highly memorable experiences, which present an auxiliary avenue in complement to optimizer improvements. 

\subsection{Uncertainty-Aware Continual Finetuning on Diverse Reasoning tasks}\label{sec:exp_reasoning}
Finally, we continually fine-tune a pre-trained Qwen3-1.7B model \citep{qwen3technicalreport} on three sequential reasoning tasks: mathematical problem solving (GSM8K) \citep{gsm8k}, code generation (Alpaca-Code) \citep{codealpaca}, and commonsense reasoning (ARC-Challenge) \citep{arc}, in this order. In addition to the performance of the new tasks (gauging plasticity), we evaluate the stability of \method across these tasks by reporting the ability of the model to solve math problems (the initial fine-tuning set) after continuing training in coding skills and common-sense reasoning (visualized in the form of shading of the respective task's color in \Cref{fig:LLM_task}). As the previous experimental subsections have already corroborated the positive impact of the slow-update posterior merging, we directly evaluate the full \method variant only. 

\Cref{fig:LLM_task} illustrates that \method achieves comparable plasticity on each new reasoning task to standard finetuning with AdamW, matching $\sim$65\% on math problems, $\sim$52\% on code generation, and $\sim$80\% on the ARC-Challenge. When we then re-evaluate the model’s initial math-solving ability after learning these new skills, CoVON exhibits only minimal forgetting: performance drops by just 2\% after code generation and 4\% after common-sense reasoning. Similarly, \method-NoM also observes high stability across tasks, reporting only 8\% forgetting at the end of the training. In contrast, AdamW exhibits a $\sim$70\% degradation in mathematical reasoning after commonsense finetuning, and EWC$^*$ incurs additional computational overhead without matching \method's retention ($\sim$ 18\% drop). These results further highlight the importance of the posterior merging as part of the slow consolidation step. Embedded together, they show that the interplay between fast adaptation and slow consolidation makes \method a robust continual optimization method that scales to challenging tasks.

\section{Conclusion}
In this work, we have realized a natural fast and slow adaptation mechanism within a Variational Continual Learning framework and have shown that it can be instantiated in a practical optimizer for continual adaptation of a single model. In our proposed \method approach, the two timescales were implemented through task-wise posterior updates, which entail rapid adaptation to new data, and posterior merges, which entail slower, uncertainty-aware knowledge preservation across tasks. Across domain-incremental classification, continual pre-training, and fine-tuning of large language models, these mechanisms were shown to consistently improve over existing VCL optimizers and weight-regularization methods. For future work, \method's ease of use and Adam-like implementation position it favorably for combination with auxiliary structural continual learning perspectives, such as memory replay, dynamic architecture modifications, or in-context learning mechanisms \citep{dorovatas2026modularmemorykeycontinual}. One limitation of \method is that it relies on Gaussian posteriors with diagonal variance, extending it to more expressive posteriors is therefore a natural next step. Another promising avenue lies in addressing related knowledge interference challenges, e.g., in federated learning or unlearning, which we posit could be generalized by drawing inspiration from the outlined perspective of knowledge adaptation as posterior correction in \citet{khan2025knowledge}. 

\section*{Acknowledgments}
This work was supported by the Bayes duality project, JST CREST Grant Number JPMJCR211. TM acknowledges the support of JSPS KAKENHI Grant Number 26H02541. 
This research has also benefited from Germany’s Excellence Strategy EXC-3057 ``RAI: Reasonable Artificial Intelligence'', funded by the Deutsche Forschungsgemeinschaft (DFG, German Research Foundation), as well as the high-profile area “Minds, Media, Machines” (MMM) at the University of Bremen. 

\bibliographystyle{icml2025}
\bibliography{refs}

\newpage

\appendix
\crefalias{section}{appendix}
\crefname{appendix}{Appendix}{Appendices}
\onecolumn

\section{Derivation and Algorithms}
\label{sec:app_derive}
\subsection{Optimality Condition of the Variational Learning Problem}
\label{app:derivation-vclgeneral}

We start from the variational continual learning objective, 
\begin{equation*}
   q_{t}^\textrm{fast} \gets \arg \min_{q \in \mathcal{Q}} \, \lambda_{t} \mathbb{E}_q[\ell_{t}] + \dkl{}{q}{q_{t-1}^\textrm{slow}}.
\end{equation*}

We write the optimality condition of this objective, following \citet{khan2025knowledge} (see their Eq. 3), 
\begin{equation*}
    q_t^\textrm{fast}(\vparam) \propto q_{t-1}^\textrm{slow}(\vparam) \exp\left( -\lambda_t \vT(\vparam)^\top {\nabla}\myexpect_{q_t^\textrm{fast}}[-\ell_t] \right) = q_{t-1}^\textrm{slow}(\vparam) \exp( -\hat\ell_t )^{\lambda_t}, 
\end{equation*}
where ${\nabla}$ is the natural gradient, $\vT(\vparam)$ is the sufficient statistics of the exponential family distribution parameterizing $q$, and we define $\hat\ell_t(\vparam) = \vT(\vparam)^\top {\nabla}\myexpect_{q_t^\textrm{fast}}[-\ell_t]$ to ease notation. Further details for this derivation are in \citet{khan2025knowledge}. We can see that this is equal to multiplying the prior $q_{t-1}^\textrm{slow}(\vparam)$ by the effect of the new data $\exp( -\hat\ell_t )$ raised to the power $\lambda_t$. 

\subsection{Optimality Condition of the Variational Learning Problem for Gaussians}
\label{app:derivation-optcond}
In this section, we perform the same derivation as in \Cref{app:derivation-vclgeneral}, but using a different approach, and specializing from the beginning to Gaussian approximate distributions. 

We start from the variational continual learning objective, 
\begin{equation*}
   q \gets \arg \min_{q \in \mathcal{Q}} \, \lambda_{t} \mathbb{E}_q[\ell_{t}] + \dkl{}{q}{q_{t-1}}.
\label{eq:vcl_app_1}
\end{equation*}
We restrict the variational family to be Gaussian, 
\[ 
q = \mathcal{N}(\vm, \vSigma). 
\]
$\vm$ denotes the parameter mean and the corresponding precision is defined as: \[ \vS=\vSigma^{-1}.
\]

We write the objective over $q$ in terms of the parametrization $(\vm, \vSigma)$ as follows:
\begin{equation*}
   \mathcal{J(\vm, \vSigma)} = \lambda_{t} \mathbb{E}_q[\ell_{t}] + \dkl{}{\mathcal{N}(\vm, \vSigma)}{\mathcal{N}(\vm_{t-1}, \vSigma_{t-1})}.
\label{eq:vcl_app_2}
\end{equation*}
The KL divergence between two Gaussians is given as, 
\begin{align*}
&\dkl{}{(\vm, \vSigma)}{(\vm_{t-1}, \vSigma_{t-1})} =\\
&\quad \frac12\Big[ \tr(\vS_{t-1}\vSigma) +
(\vm-\vm_{t-1})^\top \vS_{t-1} (\vm-\vm_{t-1}) - d + \log\det\vSigma_{t-1}
- \log\det\vSigma \Big].
\end{align*}

Discarding constants independent of \( (\vm, \vSigma)\), the remaining terms relevant for the optimization are:  
\[
 \frac12\tr(\vS_{t-1}\vSigma)
+ \frac12(\vm-\vm_{t-1})^\top\vS_{t-1}(\vm-\vm_{t-1})
- \frac12\log\det\vSigma.
\]

The derivative of the objective with respect to \(\vm\) is
\[
\nabla_{\vm}\mathcal J
=
\lambda_{t}\,\mathbb{E}_q[\nabla \ell_{t}(\vparam)]
+
\vS_{t-1}(\vm-\vm_{t-1}).
\]
Setting this gradient to zero gives
\begin{equation*}
\vS_{t-1}\big(\vm-\vm_{t-1}\big)
=
-\lambda_{t}\,\mathbb{E}_{q}[\nabla \ell_{t}].
\end{equation*} 
This is the first optimality condition shown in the main paper.

Next, differentiating with respect to the covariance, we get:  

\[
\nabla_{\vSigma}\mathbb{E}_q[\ell(\vparam)] = 
\frac12\mathbb{E}_q[\nabla^2\ell(\vparam)].\]
Together with 
\[
\nabla_{\vSigma}\frac12\tr(\vS_{t-1}\vSigma) = \frac12\vS_{t-1}, 
\qquad 
\nabla_{\vSigma}\!\left(-\frac12\log\det\vSigma\right)
=
-\frac12\vSigma^{-1},
\]
we obtain 
\[
\nabla_{\vSigma}\mathcal J
=
\frac{\lambda_{t}}{2}\mathbb{E}_q[\nabla^2 \ell_{t}(\vparam)]
+
\frac12\vS_{t-1}
-
\frac12\vSigma^{-1}.
\]

Setting this gradient to zero yields
\[
\vS
=
\vS_{t-1}
+
\lambda_{t}\,\mathbb{E}_{q}[\nabla^2 \ell_{t}].
\]
This is the second optimality condition shown in the main paper.

\subsection{Baseline Implementations of VCL}
\label{app:adareg}
We show that VCL can be approximately implemented with standard first-order optimizers like AdamW using its online Hessian estimates, as shown in \Cref{alg:adareg}. More specifically, \method can be instantiated as a standard first-order continual optimization algorithm, which we call AdamW-Regularized (Ada-Reg). For completeness, we also provide pseudocode for EWC and EWC$^*$~\Cref{alg:ewc-basic} as Laplace approximations to VCL, as well as variational continual learning implemented with Bayes-by-Backprop~\Cref{alg:vcl}. 

We emphasize that EWC and Ada-Reg arise as a degenerate special case of \method under the following assumptions:
\begin{enumerate}
\item We approximate posterior expectations at the mean (no sampling).
\item For EWC and EWC$^*$, curvature is estimated once per task using squared gradients and then frozen. For Ada-Reg, Hessian curvature is approximated using the squared-gradient accumulator of Adam. 
\item Optimization proceeds deterministically using only the posterior mean. 
\end{enumerate}
\method relaxes all three assumptions, yielding a strictly more general optimizer that retains uncertainty awareness while remaining cheap for large-scale training. 
\begin{algorithm*}[t]
\caption{AdamW-Regularized (Ada-Reg)}\label{alg:adareg}
\begin{algorithmic}[1]
\Init{
$\vparam \gets (\text{NN-weight init}), \;
\vparam_0 \gets 0, \;
\vs_0 \gets \text{prior-precision}$
}
\Init $\beta_1, \beta_2 \in (0,1)$,  $\lambda_t$ for each task $t$, Hessian init $\vh_0 = 0$, \text{learning rate} $\alpha$ 
\For{$t=1\hdots T$} 
\State $\vg \gets 0$, $\vh \gets h_0$,  \quad 
\For{$i=1,2,\hdots$}   
\State $\widehat \vg \gets \widehat \nabla \ell_{t}(\vparam)  $
\State $\widehat \vh \gets \vg^2$ 
\State $\vg \gets \beta_1 \vg + (1-\beta_1) \widehat \vg$, \quad $\bar \vg \gets \vg / (1 - \beta_1^i)$
\State $\vh \gets \beta_2 \vh + (1-\beta_2) \widehat \vh$, \quad $\bar \vh \gets \vh / (1 - \beta_2^i)$
\State $\vparam \gets \vparam - \alpha (\bar \vg / ( \sqrt{\bar \vh} + \epsilon) + \lambda_{t}^{-1} \vs_{t-1} (\vparam - \vparam_{t-1}))$
\EndFor 
\State $\vs_{t} \gets \vs_{t-1} + \lambda_{t} \bar \vh$
\State $\vparam_t \gets \vparam$
\EndFor
\end{algorithmic}  
\label{alg:vcl-ivon}
\end{algorithm*}

\begin{algorithm}[t!]
\caption{EWC and EWC$^*$ using post-hoc uncertainty with squared gradients}
\begin{algorithmic}[1]
\Init $\vparam \gets (\text{NN-weight init})$, $\vparam_0 \gets 0$, $\vs_0 \gets \text{prior-precision}$
\Init Weightings $\lambda_t$ for each task $t$
\For{$t=1\hdots T$}

\For{$i=1,2,\hdots$} 
\State $\widehat \vg \gets \widehat \nabla \ell_t(\vparam)$
\State Update $\vparam$ by optimizer step on $\widehat \vg$ and $\sum_{k=0}^{t-1} \lambda_k^{-1} \vs_k (\vparam - \vparam_k)$ (decoupled in AdamW)
\EndFor
\State $\vh \gets 0$, $k \gets |\data_t| / B$
\For{each minibatch of size $B$ in $\mathcal{D}_t$}
\State $\vh \gets \vh + \widehat \nabla \ell_t(\vparam)^2 / k$
\EndFor 
\State $\vs_{t} \gets B \lambda_t \vh$ for EWC, \quad  $\vs_{t} \gets \vs_{t-1} + B \lambda_t \vh$ for EWC$^*$
\State $\vparam_{t} \gets \vparam$, 
\EndFor
\end{algorithmic}
\label{alg:ewc-basic}
\end{algorithm}

\begin{algorithm}[t!]
\caption{Variational Continual Learning (VCL) with Bayes by Backprop \citep{nguyen2017variational}}
\begin{algorithmic}[1]

\Init Initial posterior $q(\vparam) \gets p_0(\vparam)$, prior $p_0(\vparam) \gets \gauss(\vparam\,|\, 0, \vs_0^{-1})$
\Init $\lambda_t$ for each task $t$
\For{$t=1 \hdots T$}
    \For{$i=1,2,\hdots$}
            \State Update $q$ via Bayes-by-Backprop on $\lambda_t \mathbb{E}_q[\ell_{t}] + \dkl{}{q}{q_{t-1}}$  and SGD

    \EndFor
    
    \State $q_{t} \gets q$
\EndFor
\end{algorithmic}
\label{alg:vcl}
\end{algorithm}

\section{Understanding the Hyperparameters of CoVON}
In this section, we provide practical guidance for choosing the hyperparameters that govern the training dynamics of \method. In particular, the behavior of the fast inner loop and slow outer loop depends on several key choices, including Hessian initialization, effective sample size, Hessian Momentum, $\gamma$ for the posterior-merging mechanism, etc. We now analyze the sensitivity of \method to these hyperparameters and identify the regimes in which it achieves a favorable stability–plasticity trade-off. 
\begin{figure*}[t]
    \centering
    \includegraphics[width= 0.49\linewidth]{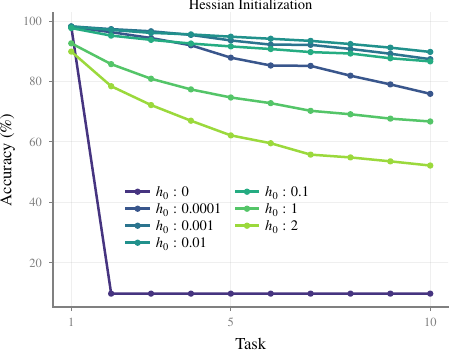}
    \includegraphics[width= 0.49\linewidth]{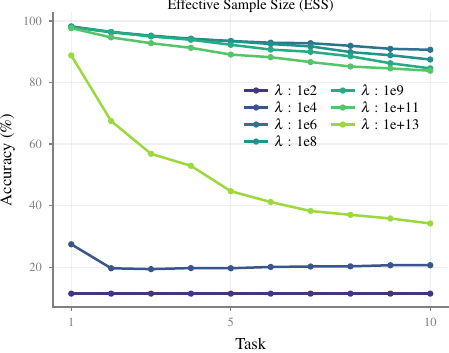}
    \includegraphics[width= 0.49\linewidth]{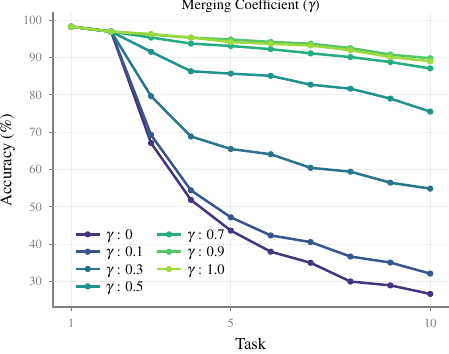}
    \includegraphics[width= 0.49\linewidth]{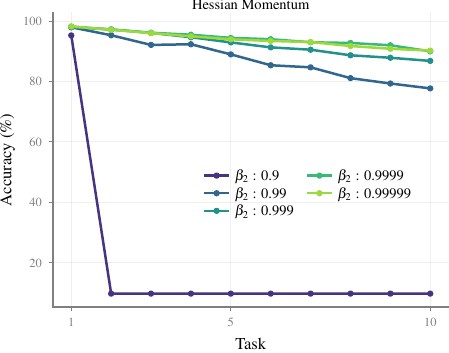}
    \caption{We analyze the importance of the different hyperparameters that influence the training dynamics of \method.}
    \label{fig:hyperparameter_analysis}
\end{figure*}

\textbf{Hessian Initialization $h_0$}: The initialization of the curvature $h_0$ defines the precision of the Gaussian prior before any data is observed.  In a continual setting, this choice is critical: an overly large $h_0$ prevents meaningful posterior formation on the initial task, whereas an overly small $h_0$ causes the initial task to dominate, leading to excessive curvature growth and premature freezing in later tasks, especially in image classification tasks. For vision tasks across different domains, values of $h_0$ in the range of $1e^{-1}$ to $1e^{-2}$ tend to work well, whereas for language models, an initialization around $1e^{-3}$ provides the best results. It is also worth noting that one needs to re-initialize the Hessian at the start of each new task. 

\textbf{Effective sample size}: Another crucial hyperparameter in \method is the effective sample size $\lambda_t$ for each task. This quantity controls both the amount of noise injected into the parameters and the influence of posterior precision in the update. In particular, the variance of the injected noise is inversely proportional to the effective sample size, ($\text{noise} \propto \frac{1}{\text{ess}}$). Larger values therefore reduce the noise and the strength of the quadratic regularization term, causing the optimizer to rely more strongly on the new data. Smaller values have the opposite effect, regularizing stronger towards the previous task.
Across both language modeling and the image classification task, we found the suitable range of ess to be around $ 10^{8}$ to $10^{10}$. Ideally, setting ess to the size of the dataset could be a good starting point.  

\textbf{Learning rate $\alpha$:} For ViT-based architectures trained on Core50, CDDB, initial learning has been set around $1e^{-5}$, except for DomainNet, it requires a higher learning rate of $1e^{-2}$. From the second task onwards, we decayed the lr in the range of $5e^{-5}$ to $1e^{-6}$ In the case of continually pretraining GPT-2 with 125M parameters, we have set the initial learning rate to be 0.97, whereas the finetuning of the Qwen3-1.7B architecture, we have set the learning rate to be $3e^{-3}$.

\textbf{Hessian momentum $\beta_2$:} The Hessian momentum needs to be set rather close to one; for instance, values in the range of 0.9999 to 0.99999 have worked well for our experiments. 

\textbf{Gradient momentum $\beta_1$:} We have set the gradient momentum $\beta_1$ to 0.9 for all of our experiments.

\textbf{Merging coefficient $\gamma$: } The $\gamma$ acts as a merging coefficient for the slow component in \method, controlling how strongly the current posterior is consolidated with the past posteriors. Across different domain-incremental classification benchmarks, a value of 0.1-0.5 worked well, whereas for the language modeling, a value in the range of 1e-2 worked well.  

\section{Additional Training and Experimental Details}\label{app:exp_details}
We now provide additional experimental details detailing different choices of hyperparameters. Our code is available at \url{https://github.com/paulsubarna/Continual-IVON/tree/main}
\subsection{Comparing against the SOTA VCL benchmark}

We follow the same experimental protocol as \citet{TDVCL} and evaluate on the Permuted MNIST Hard benchmark. The benchmark consists of 10 sequential tasks, each with 50k training samples and 10k test samples. All methods are trained with a simple MLP architecture with two hidden layers [100,100] and a single head with 10 units, and the baseline methods use the same hyperparameters reported in \citep{TDVCL}. For CoVON, we use a learning rate of 0.02 on the first task and reduce it to 0.003 for all subsequent tasks, training for 30 epochs on each task. The Hessian is initialized to 0.01, while $\lambda$ and $\gamma$ are chosen in the ranges of $1e^7-1e^8$ and 0.8 - 0.9, respectively, across the tasks.

\subsection{Contextualizing \method In Domain Incremental Settings}
We consider three standard domain-incremental benchmarks: DomainNet \citep{DomainNet}, CDDB \citep{CDDB}, and CORe50 \citep{core50}. For the fairness of comparison, we follow the same experimental protocol as prior baselines \citep{PINA}, using a ViT-B/16 backbone pretrained on ImageNet-1K. For each dataset, we use the official train/test splits with the hyperparameters used in the prior works. The corresponding CoVON hyperparameters are as follows: an initial Hessian value of $5e^{-3}$, with the Hessian for subsequent tasks set to 1e-3, $\lambda$: 1e8, $\gamma$: 0.3, and a learning rate of $1e^{-4}$ and decayed to $1e^{-6}$ from the second task onwards. 

\subsection{Continual Pretraining on multilingual data} 
For continual pretraining of the language model, we use a GPT-style architecture with \(n_{\mathrm{layer}}=12\), \(n_{\mathrm{head}}=12\), and embedding dimension \(n_{\mathrm{embd}}=768\), with dropout set to \(0.0\) and no bias terms. Training is performed with micro-batch size \(20\), gradient accumulation steps \(24\), and sequence length \(1024\), yielding a larger effective batch through accumulation. We use \(15000\) optimization steps in total, corresponding to \(5000\) updates per language for the sequence \(\{\mathrm{en}, \mathrm{de}, \mathrm{fr}\}\).

For the \method-specific hyperparameters, we use learning rate \(9.72 \times 10^{-1}\), weight decay \(1 \times 10^{-6}\), \(\beta_1 = 0.9\), \(\beta_2 = 0.99995\), gradient clipping \(1.0\), and Hessian update value \(h_{\mathrm{ess},2} = 1.92 \times 10^{-2}\). Learning-rate decay is enabled with \(500\) warmup iterations, decay over the full \(15000\) iterations, and minimum learning rate \(0\).
We set the initial Hessian to \(1 \times 10^{-3}\), the effective sample size to \(1 \times 10^{10}\) for English, \(1.44 \times 10^{13}\) for German, and \(1 \times 10^{15}\) for French, and the clipping radius to \(3 \times 10^{-4}\). The slow-update coefficients are \(\gamma = 10^{-2}\) for English, \(\gamma = 10^{-3}\) for German, and \(\gamma = 10^{-2}\) for French.

For the baselines trained with AdamW, we use a learning rate \(6 \times 10^{-4}\), a weight decay \(1 \times 10^{-1}\), \(\beta_1 = 0.9\), \(\beta_2 = 0.95\), and gradient clipping \(1.0\). Training runs for \(15000\) iterations, with learning-rate decay enabled, \(500\) warmup iterations, decay over the full \(5000\) iterations per language, and a minimum learning rate of \(6 \times 10^{-5}\).
\subsection{Continual finetuning of Qwen models on a set of sequential reasoning tasks }

We provide additional details on the training and evaluation pipeline used in our continual fine-tuning experiments. In this setting, we fine-tune a pretrained Qwen3-1.7B model \citep{qwen3technicalreport} sequentially on three reasoning tasks: mathematical reasoning, code generation, and common-sense reasoning. The datasets are GSM8K, without additional augmentation, with 7.8k training examples and 1.3k test examples; Alpaca Code, with roughly 20k training samples; and ARC-Challenge, which we use as the common-sense reasoning task, with 7.8k question–answer pairs. For evaluation, we use lm-eval-harness \citep{eval-harness} as a unified benchmark pipeline. We evaluate its test sets and set the generation length to 1024 tokens. 

The math dataset is trained for \(3\) epochs with learning rate \(3 \times 10^{-3}\), batch size \(4\), gradient accumulation \(8\), and maximum sequence length \(512\). The code stage is trained for \(2\) epochs with learning rate \(3 \times 10^{-1}\), batch size \(2\), gradient accumulation \(16\), and maximum sequence length \(1024\). The ARC stage is trained for \(3\) epochs with learning rate \(3 \times 10^{-1}\), batch size \(4\), gradient accumulation \(8\), and maximum sequence length \(256\).
We set the initial Hessian to \(1 \times 10^{-1}\), the effective sample size to \(1 \times 10^{9}\), the clipping radius to \(1 \times 10^{-2}\), and the momentum parameters to \(\beta_1 = 0.9\) and \(\beta_2 = 0.99995\) on the first task. At the transition to the code task, we set hess to \( 5 \times 10^{-1}\) and \(\lambda = 1 \times 10^{7}\). At the transition to the ARC task, we again set hess to \( 5 \times 10^{-1}\), with \(\lambda = 1 \times 10^{6}\).
Across all stages, we use evaluation batch size \(4\) and warmup ratio \(0.03\)

We compared \method with standard baselines such as AdamW and IVON finetuning (AdamW-FT / IVON-FT), Ada-Reg, a deterministic variant of CoVON, and finally regularization techniques such as EWC$\*$. In each column of \Cref{tab:ft_llm_reason}, we report the performance of the model on each new task and its ability to retain the ability to solve math tasks after learning a new skill. \method shows the best retention ability (61\%) after it has been finetuned on common-sense reasoning skills. Note, we don't use any replay for any of our methods and only opted for pure Supervised Fine-Tuning. It establishes \method as a robust and scalable optimization algorithm for LLM models.

\textbf{Compute Resources: }All experiments were conducted on four NVIDIA A600 GPUs, each with 24,GB of memory. A single run on DomainNet, CDDB, and CORe50 took approximately 300, 120, and 120 minutes, respectively. For language-model experiments, continual pretraining required about 900 minutes per run, while continual fine-tuning took approximately 100 minutes.

\begin{table*}[t]                                     \caption{Continual Fine-tuning of Qwen3-1.7B-Base on Math (\textbf{GSM8K}) $\rightarrow$ Code (\textbf{Code-Alpaca 20K})$\rightarrow$   Common-sense reasoning (\textbf{Ai2$\_$arc}). We specifically evaluate \method's ability to retain mathematical
  problem-solving capabilities after the model is sequentially fine-tuned on code generation
   and common-sense question-answering tasks. Each column reports performance on the new task and on the math task. We observe that AdamW exhibits significantly higher forgetting than \method. After fine-tuning on the code, AdamW suffers a nearly 30\% drop in math performance, compared to only 3\% for \method. A similar trend is observed after finetuning on the common-sense dataset thereafter.  }                                
      \label{tab:ft_llm_reason}                                      
      \centering                           
            \begin{tabular}{c|c c c}                                       
           \multirow{2}{*}{Methods} & \textbf{Task 1}: Math (Exact-Match $\uparrow$) & \textbf{Task 2:} Code (Pass@1 $\uparrow$) &
  \textbf{Task 3:} Arc (Acc $\uparrow$) \\  
           & \text{Eval: Math} & \text{Eval: Code/Math} & \text{Eval: Arc/Math} \\
           \hline                                                 \multicolumn{4}{c}{\emph{Supervised Finetuning (SFT) of the full model (Qwen3-1.7B-Base)}} \\                                                                               
           \hline 
           AdamW-FT  & 64.5 &  52.9/56.4  &  79.8/20.4  \\ 
           IVON-FT & 64.7 & 52.9/56.5 & 79.9/20.8 \\
           Ada-Reg & 64.5 & 52.9/58.1 & 79.8/52.2\\
           EWC* & 64.5 & 52.9/59.1 & 79.8/52.9 \\
           \rowcolor[gray]{0.92}
           \method   & 64.7 &  52.9/62.3  &  79.9/61.2  \\     
               
           \hline       
      \end{tabular}    
      \vspace{4pt}
  
  \end{table*}

\begin{table*}[t]
  \centering
  \label{fig:cost_comp}
      \caption{Comparison of accuracy, runtime overhead, and GPU memory usage in the domain-incremental setting. We report the final averaged accuracy together with wall-clock overhead measured per batch, per epoch, and per task, as well as peak GPU memory consumption in GB. While EWC and EWC$^*$ require substantial task-level overhead due to additional post-hoc computations, \method remains nearly as efficient as standard fine-tuning and Ada-Reg, with only marginal increases in per-batch and per-epoch time and no meaningful extra memory cost. Despite this comparable computational footprint, \method achieves the highest average accuracy, demonstrating a substantially better efficiency--performance trade-off than existing regularization-based baselines.}

\begin{tabular}{lccccc}
\toprule
\textbf{Method} & \textbf{Averaged}  & \multicolumn{4}{c}{\textbf{ Overhead (sec)}} \\
 &  \textbf{Accuracy}  & \textbf{ batch} & \textbf{ epoch} & \textbf{task} & \textbf{GPU Memory}\\
\midrule
Finetune       & 28.48  & 0.89s & 31.89s & 490s & 16.13 (GB)  \\
EWC            & 22.46   & 0.90s & 32.5s & 1956s & 16.13 (GB)   \\
EWC$^*$ & 36.45   & 0.90s & 32.5s & 1789s & 16.13 (GB) \\
Ada-Reg       &  49.75  & 0.92s   & 36.0s & 534s & 16.57 (GB)\\
\method      & 56.75  & 0.93s   & 36.5s & 537s & 16.57 (GB)\\
\bottomrule
\end{tabular}

\end{table*}

\subsection{Computational Overhead and Memory Cost}
In \Cref{fig:cost_comp}, we compare the runtime and memory cost of \method against standard continual learning baselines in the domain-incremental setting. \method achieves the highest average accuracy of 56.75\%, improving over Ada-Reg (49.75\%) and outperforming finetuning (28.48\%) and EWC-based methods. At the same time, its computational overhead remains very small: the per-batch time is 0.93s and the per-epoch time is 36.5s, which are nearly identical to AdamW-Reg (0.92s and 36.0s) and only slightly above plain finetuning (0.89s and 31.89s). Its total per-task cost is 537s, again very close to AdamW-Reg (534s), while EWC and EWC-Corrected require much larger task-level overheads of 1956s and 1789s, respectively, due to additional post-hoc computations. In terms of memory, \method uses 16.57 GB of GPU memory, matching AdamW-Reg and only slightly exceeding finetuning and EWC (16.13 GB). Overall, the table shows that \method improves accuracy without introducing meaningful runtime or memory overhead, providing a substantially better efficiency–performance trade-off than existing regularization-based baselines. 
\label{app:baselines}


\end{document}